\documentclass[10pt,twoside,twocolumn,a4paper]{article}

\usepackage[accepted]{bpasts}

\usepackage{t1enc}
\usepackage[utf8]{inputenc}
\usepackage{amsmath,amsfonts}

\usepackage{graphicx}
\usepackage{flushend}

\abtitle{MCTS for The Lord of the Rings}
\title{Optimisation of MCTS Player for The Lord of the Rings: The Card Game}

\abauthor{K. Godlewski, B. Sawicki}
\author{K. GODLEWSKI\email{jowisz11@gmail.com} , B. SAWICKI \email{bartosz.sawicki@pw.edu.pl}}

\Address{Warsaw University of Technology, ul. Koszykowa 75, 00-662 Warsaw, Poland}

\Abstract{The article presents research on the use of Monte-Carlo Tree Search (MCTS) methods to create an artificial player for the popular card game "The Lord of the Rings". The game is characterized by complicated rules, multi-stage round construction, and a high level of randomness. The described study found that the best probability of a win is received for a strategy combining expert knowledge-based agents with MCTS agents at different decision stages. It is also beneficial to replace random playouts with playouts using expert knowledge. The results of the final experiments indicate that the relative effectiveness of the developed solution grows as the difficulty of the game increases.}


\vol{XX} \no{Y} \year{2016}
\doi{10.1515/bpasts-2016-00ZZ}

\begin{document}
\maketitle


\section{Introduction}

"The Lord of the Rings: The Card Game" is one of the most popular card games. Since its launch in 2011 by Fantasy Flight Games, it has gained great popularity, as evidenced by more than 100 official expansions, dozens of internet blogs, and millions of fans worldwide. This game's uniqueness and enormous success are due to its cooperative nature and the fact that it can also be played in solo mode. By default, the core set of cards supports up to 2 players, but the game can be played by 3 or 4 players with additional core sets. The players have to fight against the deck of cards representing the forces of Sauron, which are obstacles to overcome.  To the best of the authors' knowledge, the game has not yet received an AI player able to win at a level comparable to human experts.

The Monte-Carlo tree search (MCTS) is a stochastic algorithm, which proved its unique power in 2016 by beating the human master in the Go game. It was the last moment when human players had a chance to compete with AI players. Since that time there is growing number of applications of MCTS in various games~\cite{Browne2012},\cite{b5},\cite{b7}. 
The MCTS as a general-purpose heuristic decision algorithm also has many applications outside the world of games. These include combinatorial optimization, scheduling tasks, sample-based planning, and procedural content generation~\cite{Browne2012},\cite{Mandziuk2018}. More recently, domain of usage has been expanded to material design~\cite{Kajita2020}, network optimisation~\cite{Haeri2017}, multi-robot active perception~\cite{Best2020}, cryptography~\cite{dhar2018} and others \cite{guzek2011}, \cite{tefelski2013}.

In this paper, MCTS algorithm is successfully used in the cooperative card game, which could be treated as a novelty comparing to the studies~\cite{Ward2009} of classical competitive games such as Magic: The Gathering.

This is especially important nowadays when cooperative games gain new applications. They have demonstrated that they are an effective, modern educational tool~\cite{Turkay2012}, e.g., in medicine~\cite{Bochennek2007}. Cooperation allows to learn teamwork, which is often not possible without the additional presence of artificial players with decision-making skills. 

An interesting approach for MCTS application in collectible card games was presented in 2019 using the "Hearthstone" game~\cite{Choe2019}. The authors identified the huge size of the action space. Several precautions were taken to reduce the number of allowed moves. Using Action filtering and Obliged Actions are examples of successful domain-specific knowledge incorporation.

This article is directly based on the authors' conference work~\cite{Godlewski2020}. The algorithm has been extended by adding more expert knowledge into the standard MCTS implementation. This
allowed to perform a new analysis for games with high complexity
level. We have demonstrated that the relative effectiveness
of the mixed strategy proposed in~\cite{Godlewski2020} rises as the
difficulty of the problem increases. 

\section{The Lord of the Rings: The Card Game}

"The Lord of the Rings: The Card Game", often abbreviated as LoTR, is a complex cooperative card game with several decision-making stages.  The following section is devoted to the presentation of its basic rules. It is necessary to understand before we will discuss the construction of the game simulator and searching for the optimal strategy for the AI player.

\subsection{Living Card Game}

From Poker to "Magic: the Gathering", card games belong to a group of games characterized by hidden information and a high degree of randomness. Hidden information means that the player does not have a complete view of the game, opposing cards and cards in the deck are unknown. At certain moments of the game, he has to draw a card - then there is a random factor with dynamically changing probabilities of drawing a certain card, depending on the previous state of the game. In the wide world of card games, in addition to the classic variants based on a 52-element stack, there are also systems with a much larger number of cards, where the player has the option of "deckbuilding" - he makes his own stack from all available cards. The cards have statistics such as hit points, attack/defense and mini-scripts that have a specific effect in the game. These types of card games can be divided into Living Card Games (LCG), in which the player expands his deck by purchasing expansions with strictly defined cards and scenarios, and Collectible Card Games (CCG), in which the purchased packs contain random cards, which gives a certain unpredictability to the whole "deckbuilding" process.

The most famous CCG game is undoubtedly "Magic: The Gathering", which was released in 1992~\cite{b1}. Players use their decks to duel with each other, the goal is to reach 20 points. LCG games such as "Star Wars", "Game of Thrones" and finally "Lord of The Rings" (LoTR) have enjoyed increasing popularity since the 2000s based on movies or series. Cooperation is a unique feature of LoTR, where players (from 1 to 4) work together to defeat all opponents and finish the scenario.

\subsection{Rules of the game}

In LoTR, the player's task is to complete the scenario, consisting of three \textit{quest cards}. Each \textit{quest card} has a specific number of \textit{progress points} that must be obtained to complete a given stage of the scenario. The player receives \textit{progress points} by playing cards from his hand and then assigning them to the expedition. Opponents drawn randomly from \textit{encounter deck} hinder the progress of the expedition, additionally, in the defensive phase, they attack the player, dealing damage to hero and ally cards. In addition to opponents, heroes and allies, there are other types of cards in the game: places, events, and items. Places are destination cards, where the player can travel to; events can be drawn from \textit{encounter deck}, they affect the player; items are kind of attachments to heroes, giving them buffs.

\begin{figure}[htbp]
    \centerline{\includegraphics[width=0.33\textwidth]{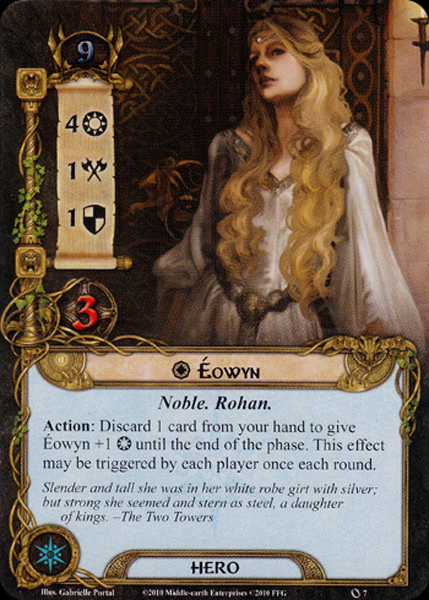}}
    \caption{Example of Hero Card: Eowyn, where 9 is Threat Cost, 4 is Willpower Strength, Attack and Defence Strength are both equal 1. The figure reproduced from~\cite{Eowyn}.}.
    \label{fig:eowyn}
\end{figure}

As seen in Fig.~\ref{fig:eowyn}, each character in the game has statistics such as \textit{hitpoints}, \textit{attack}, \textit{defense} and if as a result of the fight \textit{hitpoints} it falls below zero, then the card is discarded from the game. If the player loses all three heroes, the game ends. The statistics of \textit{willpower} and \textit {threat} are responsible for the progress of the expedition (\textit{quest resolution}), if the difference between \textit{willpower} characters assigned to the expedition and \textit{threat} of opponents and places is greater than zero, then the player places \textit{progress points} on the current scenario \textit{quest card}. When all three \textit{quest cards} are completed, the game is deemed a win. If the difference is less than zero, the player increases his threat level by this amount, exceeding the threat level over 50 means losing the game.

Every hero card has \textit{resource pool}, which is increased by one token every round, this process occurs at the \textit{Resource} stage.

The characters are grouped in four "spheres": Spirit, Tactics, Lore and Leadership. The symbol of the "sphere" is depicted on the left-bottom corner of the card. Eowyn Hero, as seen in Fig.~\ref{fig:eowyn} belongs to the Spirit sphere, which is indicated by the blue star symbol.

Each card, in addition to statistics, also has \textit{game text} - special skills invoked during the game. Player-controlled characters get buffs, whereas enemies or events cast negative effects.

\subsection{Round structure}

Before starting the game, all decks: \textit{player deck} and \textit{encounter deck} are shuffled. The player sets his initial threat level, which is the sum of the appropriate parameters of all three heroes, then draws six cards into his hand from the \textit{player's deck}. 

\begin{figure}[htbp]
    \centerline{\includegraphics[width=140pt]{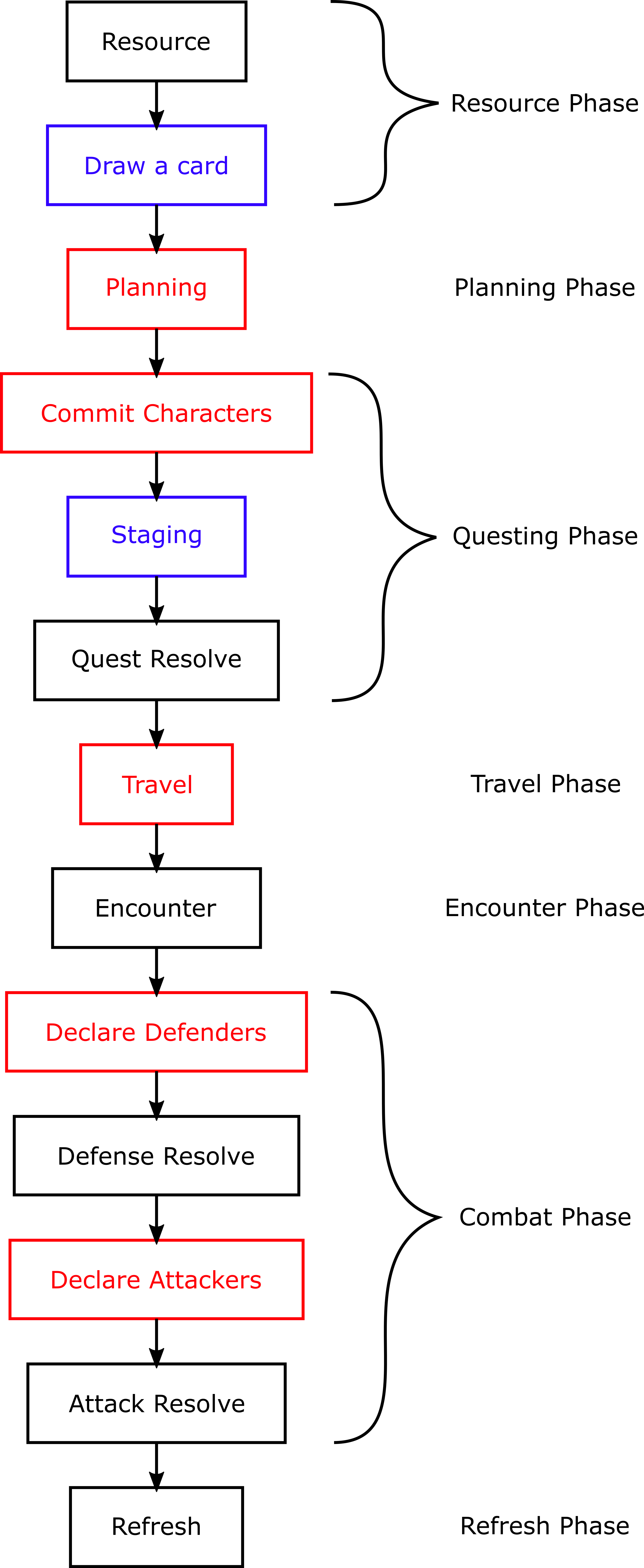}}
    \caption{Round structure: random (violet), decision (red), ruled (black) stages.}
    \label{fig:turn_structure}
\end{figure}

A diagram of one round of the game is shown in Fig.~\ref{fig:turn_structure}. There are 13 stages divided into 7 round phases. Six of these stages are a simple implementation of the described game rules (marked in black), two of them contain unpredictable random actions (violet), while the other five are stages (red) in which the player must decide. To create an AI agent, it is important to understand each decision stage~\cite{Powley2014}. 

In the \textit{Resource} stage, one token is added to \textit{resource pool} for each hero. Afterward, the player draws one card from his deck.

In the \textit{Planning} stage/phase, the player plays cards from his hand paying tokens from \textit{resource pool} of the heroes. He can only pay for a card from a "sphere", which hero belongs to, for example, hero \textit{Eowyn} belongs to the \textit{Spirit} sphere.

The \textit{Commit Characters} stage is based on assigning heroes and previously played cards for the trip through \textit{tapping} (rotating the card 90 degrees). Only the \textit{willpower} card parameter is relevant in this phase, the player determines its sum. Then the top card is taken from \textit{encounter deck} and it goes to \textit{staging area}. The sum of threat cards in this zone determines the level of adversity, which is subtracted from the sum of \textit{willpower}. At this stage of the game, randomness plays a big role, the player must allocate their cards without knowing what will fall out of \textit{encounter deck}.

In \textit{Travel} stage/phase, the player decides to go to a given place, then take the \textit{location} card from staging area and place it on the current \textit{quest card}. From this point, every \textit{progress points} obtained in \textit{questing} in future turns goes to this \textit{location}, until all \textit{quest points} are filled. Then it is removed and the player can declare another card as his destination. 

In the \textit{Encounter} phase, \textit{engagement checks} are performed: the player takes from the \textit{staging area} all opponents who have less or equal \textit{engagement cost} from the player's threat level. Then they go to \textit{engagement area}.

Next in \textit{Combat} phase, the opponents from \textit{engagement area} attack the player one after the other. He has the opportunity to \textit{Declare defenders} or to take an unprotected attack. In the first case, he turns the untapped cards and resolves the fight: the value of \textit{defense} of the defender minus \textit{attack} of the attacker, if the difference is negative, the defender loses the corresponding \textit{hitpoints}. An unprotected attack only goes to heroes, in this case their \textit{defense} is omitted. If a player has any untapped cards, he can make strike back by \textit{Declare Attackers}: select any opponent in \textit{engagement area} and resolve the fight in a similar way.

Closing the turn (\textit{Refresh} phase) is to remove all characters whose hitpoints have fallen below zero, untap cards and, finally, to increase the player's threat level by one.

\section{Game simulator}

Before starting research on computational intelligence methods in the LoTR game, it was required to create a computer simulator of the game. The developed software enables multiple, quick-playing and experimenting with different agent configurations.

\begin{figure}[htbp]
    \centerline{\includegraphics[width=240pt]{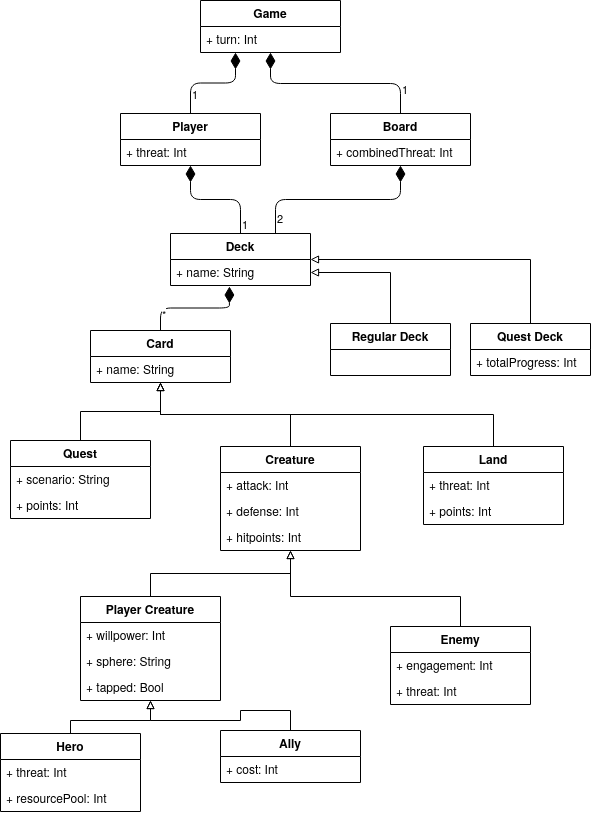}}
    \caption{Class diagram of the LoTR game simulator}
    \label{fig:classDiagram}
\end{figure}

The class structure of the LoTR game simulator (Fig. \ref{fig:classDiagram}) was designed with respect to an object-oriented paradigm in the Python programming language. The class Game includes the Player and the Board, which feature corresponding decks of cards. The inheritance of classes provides functionality separation making the code easy to extend and debug. Moreover, the game simulator is hermetic from the AI agent's point of view, therefore it is suitable for algorithm testing. Code of the simulator is available on the Github repository\footnote{https://github.com/kondziug/LotR\_simulator}.

\begin{table}
\caption{Parameters to control game complexity level}
\begin{center}
\begin{tabular}{|l|c|c|c|}
\hline
 & Easy & Medium & Hard \\
\hline
Scenario setup & no & no & yes \\  
\hline
Number of cards & 28 & 29 & 29 \\  
\hline
Number of card types & 7 & 15 & 15 \\  
\hline
\end{tabular}
\label{tbl:complexity}
\end{center}
\end{table}

The main program creates a game root and sets it up according to the difficulty level and playout budget. Each time decision is taken, the game root moves down from the current state to the new node. Once the node, which the root is being transferred to, turns out to be an ending state, the main program returns the game lose or win. 

The difficulty level specifies what types of cards (enemies or location cards) will form part of the deck. Three levels of difficulty are provided: easy, medium and hard (see~Table~\ref{tbl:complexity}). At easy level, the encounter deck consists of player-friendly enemies: their stats like hitpoints or attack-points are relatively low, whereas medium and hard levels take a full set of cards~\cite{b3}. The key difference between these two is the scenario set up - within hard level before game start certain cards are already added to the staging area: \textit{Forest Spider} and \textit{Old Forest Road}. As shown in the experiments, it poses a huge challenge for both of the MCTS methods to struggle with.

The playout budget determines how many times a game starting from a node should be rolled out to the end state. The end state can be considered as a player's win or lose according to the rules. The playout budget could also be considered as an external constrain of the algorithm. In MCTS higher budget leads to a longer time to decision and larger memory usage.

Due to the high complexity of the original game, several constraints have been applied to the simulator. Event and item cards have not been implemented because they are not essential for the whole gameplay. We also skipped special effects described on the \textit{game text}. Huge variety of those effects makes them hard to serialize in the code. To have shorter games, the scenario has been limited to only one \textit{quest card}, so the players need to get a smaller number of progress points to win.

\section{Agent players}

The steps from Fig.~\ref{fig:turn_structure} come sequentially one after the other and the options for action in a game node depend strictly on the decision made before. In the Commitment stage, the player gives up characters for the quest, he can choose one or more Heroes or characters that have already been bought during the planning. Usually, the player plays 1-2 cards from his hand, so that he forms a subset of a group of 5 cards for the commitment. The size of the subset depends on the current total threat of the cards in the stage area - one can be zero, another round 3 for example. The number of enemies in the encounter area determines the number of declared defenders - if there are 3 enemies, for example, the player must assign 3 out of 5 characters, so this makes 10 subsets. In summary, the size of the action area varies according to past events.

There are five decision stages within the round of the game: Planning, Commitment, Travel, Declaring Defender and finally Declare Attackers. The stages of Travel and Declaring Attackers are considered straightforward, so expert rules are all that was required. The other stages can be resolved in different ways. Decisions can be carried out by AI agents, or by a reduced rule-based player with a simple logic implemented (such as~\cite{cowling2012}). Finally, the analysis features four types of player agent: random, rule-based (hereafter called expert), simple-flat Monte-Carlo, and full MCTS methods.

\begin{description}

\item[Agent 1.] Random choice. Randomness has been constrained by the rules of the game. At the Planning stage random selection of cards in hand is checked if it is possible to play according to the game rules. At the Commit stage, the agent draws a subset of characters in play, if their total willpower is higher than the total threat of the cards in the staging area: commits it the quest, otherwise reject the subset and draw again. In Declare Defenders stage agent samples one character for every enemy in the engagement area. 

\item[Agent 2.] Expert knowledge.
The rules formulated before an experienced human player have been stored in the form of simple decision-making algorithms. Each stage of the game had a separate set of rules.

\item[Agent 3.] Flat Monte Carlo. The idea behind flat Monte-Carlo is to create only the first layer of a decision tree. The specified number of playouts is run for each child. Node with the highest number of wins is selected as the best choice.

\item[Agent 4.] MCTS. 
Monte Carlo Tree Search is an algorithm for taking optimal decisions through a sequentially built decision tree based on random sampling. MCTS consists of four steps repeated until a given playout budget is reached: 1)~Selection, 2)~Expansion, 3)~Simulation, 4)~Backpropagation.

\end{description}

\begin{figure}[htbp]
    \centerline{\includegraphics[width=0.5\textwidth]{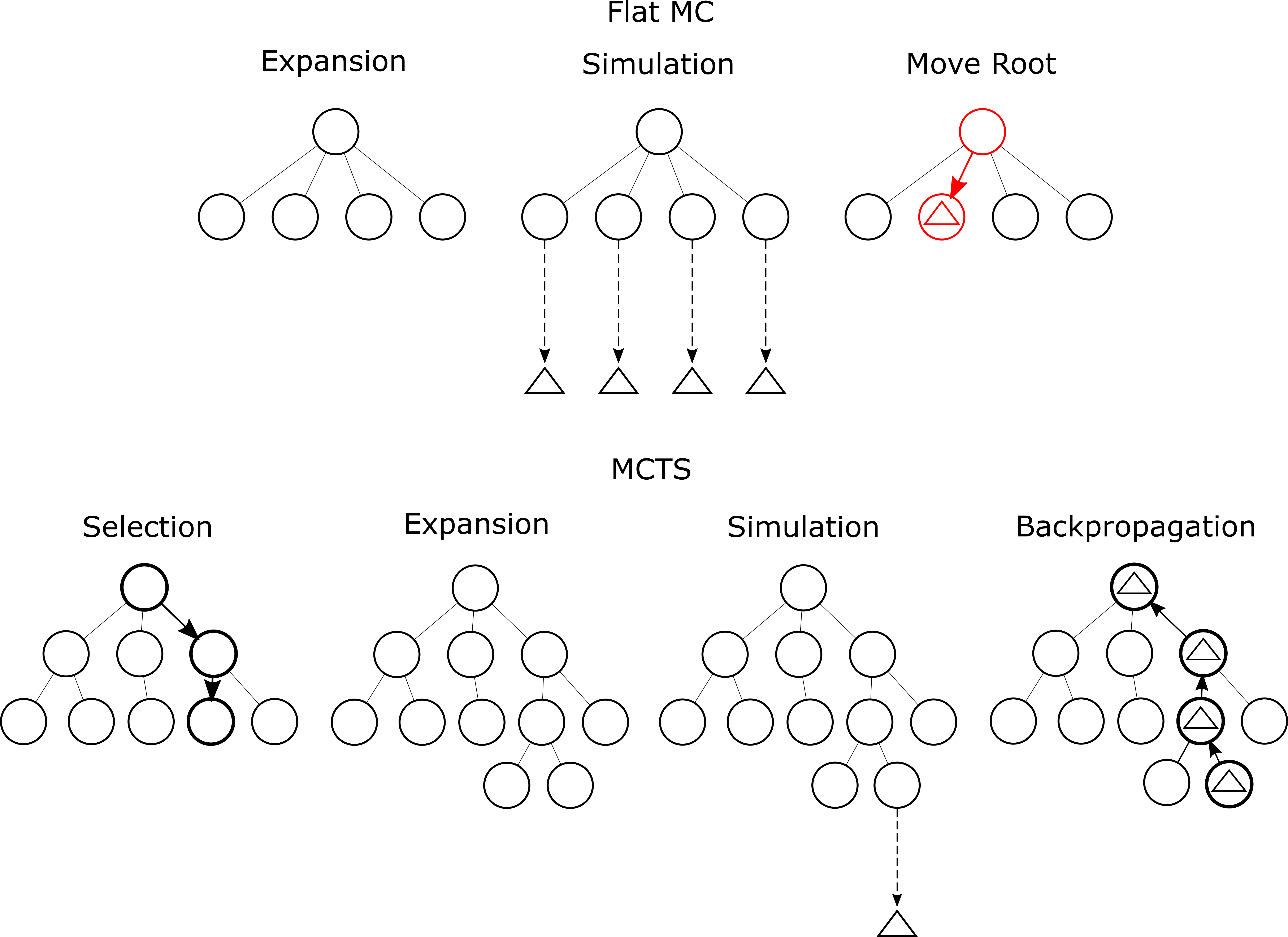}}
    \caption{Comparison of Flat Monte-Carlo, with Monte-Carlo Tree Search algorithms}
    \label{fig:MCTSsteps}
\end{figure}

The key distinction between Flat MC and MCTS is the depth of the tree (see Fig.~\ref{fig:MCTSsteps}). In Flat MC playouts are performed from every child node uniformly and the root proceeds to the node with the best winrate.
Considering only nodes from the first level of inheritance result lacks Selection and Backpropagation phases. 

In the MCTS algorithm, the state tree is as deep as limited by time or memory constraints. The main difficulty in Selection is to maintain balance between the exploitation of deep variants of moves with high winrate and the exploration of moves with few simulations. We used a classical approach based on Upper Confidence Bound for Trees (UCT)~\cite{Browne2012}:
\begin{equation}\label{eq:UCB}
x_{j} + \sqrt{\frac{2 \ln n}{n_{j}}}
\end{equation}
where $x_j$ is the current winrate for node $j$, $n$ – the number of
playouts run from the parent of node $j$, $n_j$ – the number of playouts
run from node $j$. In the Selection phase, values of UCT function is calculated for all nodes. The one with the highest value is selected.

In Expansion, new leaves are added to the selected node. It is clear that the actions under consideration must be in line with the rules of the game. For this purpose, independent validity functions have been implemented for each required decision. In the Planning stage, legal moves are determined by checking that the player's resources allow you to buy the card if you create the node; for the Questing stage all combinations of available characters are considered, if the total willpower of a given subset is greater than the total threat cards in the staging area, then a node is created.  However, it is beneficial to introduce heuristics to limit the number of new states of the game.  

Comparing with our previous results~\cite{Godlewski2020}, the use of expert rules has allowed to achieve a significant increase in the effectiveness of the method at a high level of difficulty. The extension of the algorithm is based on the limitation of analyzed actions. In the Defense stage, subsets of cards are created with correspondence to the number of opponents in the engagement area. The subsets can contain heroes already embedded in previous stages of the game. In Declare Defenders stage, a similar strategy for untapped cards and enemies in the engagement area was developed. 

In the third phase of MTCS (Simulation), playouts to the terminal state of the game are performed. We have implemented two playouts strategies: random and expert, which correspond to agents 1 and 2 in terms of implementation. 

During the Backpropagation phase (the last step of the MCTS round), statistics of the number of won playlists and the number of visits for all nodes up to root are updated. Then values of UCT functions have to be recalculated, and the next Selection phase can start.

\section{Numerical experiments}

This section describes research aimed at finding the optimal strategy of artificial players supported by computational intelligence algorithms. To properly present the statistical nature of the results, each simulation has been repeated 1000 times.

The simulations were run in parallel on the host machine with a 12-cores Intel i9-9920X processor and 128GB RAM. Spawning processes across the CPU had been executed with Python Multiprocessing Package. After pooling the results of the simulations, postprocessing was applied - for every experiment winrate with confidence interval was calculated. Binomial proportion confidence interval for 95\% confidence level is described by the equation:
\begin{equation}
    \pm z \sqrt{\frac{p (1 - p)}{n}},
\end{equation}
where $z$ - 1.96 for 95\% confidence level, $p$ - winrate probability $n_s / n$, $n$ - total number of trials, $n_s$ - number of wins.

\begin{figure}[htbp]
    \centerline{\includegraphics[width=0.5\textwidth]{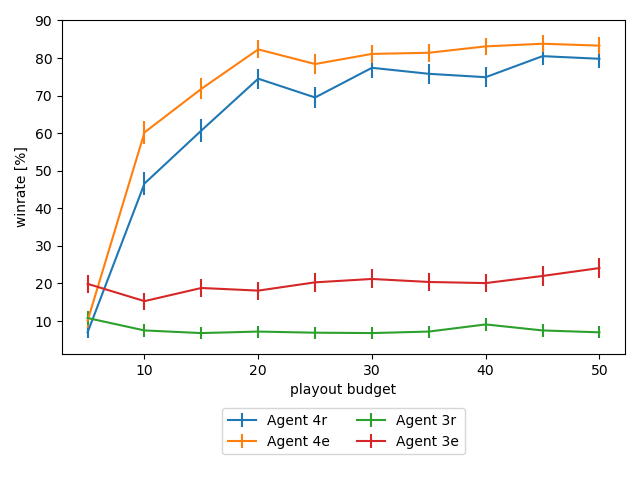}}
    \caption{Probability of winning (winrate) as a function of playout budget for different types of the playout strategies: 'r' - random, 'e' - expert. The size of statistical sample is 1000 games. Complexity level - medium.}
    \label{fig:playoutBudget}
\end{figure}

To begin with, the impact of playout budget was under investigation. Fig.~\ref{fig:playoutBudget} proves a significant advantage of MTCS (Agent 4) over Flat MC version (Agent 3). Only below 10 playout budget the performance of both agents appears to be comparable. The second interesting observation is the influence of the expert playout strategy. It emerges gradually for Agent 4, while keeping its dominance about 10-15 points since the beginning in the case of Flat MC. The saturation seen on the Agent 4 plots clearly suggests that increasing the playout budget over 40 is redundant, therefore this value will be used in further considerations.

\begin{figure}[htbp]
    \centerline{\includegraphics[width=0.5\textwidth]{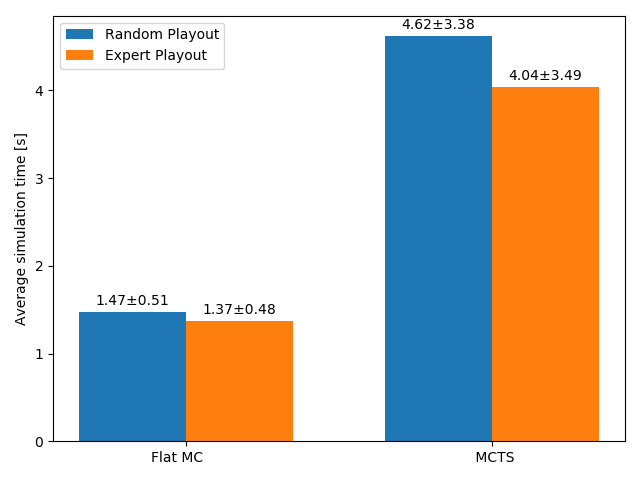}}
    \caption{Average simulation time for Agent 3 and Agent 4 for different types of playouts. Playout budget is set to 40 and complexity level to medium.}
    \label{fig:playoutTime}
\end{figure}

When it was noticed that playouts with expert knowledge have higher chances for a win, we raised a question about its computational cost. Results of measured simulation time are presented in Fig.~\ref{fig:playoutTime}. For both types of agents (Flat MC and MCTS) playout strategy does not affect the simulation time. The average time is nearly the same, when you note standard deviation reaching over 70\%. Another observation is that the Agent 4 is significantly more time-consuming, however, the volatility of simulation time is also greater. 

These two experiments lead to a conclusion that the optimal setup for playout strategy is: 40 repetitions in the budget at expert mode. Within these circumstances the methods reach a sufficient winrate at acceptable simulation time. 

These two experiments conclude that the optimal setup for the playout strategy is: 40 repetitions in budget and operation in expert mode. Under these circumstances, the methods achieve a sufficient winrate with acceptable simulation time.

\begin{table}
\caption{Winrate for combination of agents on three different decision stages (complexity level - medium, number of trials – 1000)}
\begin{center}
\begin{tabular}{|c|c|r|}
\hline
\textbf{Planning - Questing - Defense} & \textbf{Winrate} \\
\hline
agent3 - agent2 - agent2 & 98.1 $\pm$ 0.85  \\
\hline
agent4 - agent2 - agent2 &  97.1 $\pm$ 1.04 \\
\hline
agent4 - agent2 - agent4 & 96.4 $\pm$ 1.15 \\
\hline
agent2 - agent2 - agent4 &  92.8 $\pm$ 1.60 \\
\hline
agent3 - agent3 - agent2 & 82.5 $\pm$ 2.36  \\
\hline
agent2 - agent3 - agent2 & 81.6 $\pm$ 2.40   \\
\hline
agent4 - agent4 - agent4 &  80.2 $\pm$ 2.47 \\
\hline
agent4 - agent4 - agent2 & 76.5 $\pm$ 2.63 \\
\hline
agent2 - agent4 - agent4 &  67.6 $\pm$ 2.90 \\
\hline
agent2 - agent4 - agent2 &  40.2 $\pm$ 3.04  \\
\hline
agent3 - agent2 - agent3 & 39.5 $\pm$ 3.03  \\
\hline
agent2 - agent2 - agent3 & 34.8 $\pm$ 2.95  \\
\hline
agent2 - agent3 - agent3 & 24.3 $\pm$ 2.66  \\
\hline
agent3 - agent3 - agent3 & 20.4 $\pm$ 2.50 \\
\hline
\end{tabular}
\label{table:1}
\end{center}
\end{table}

The next investigated problem was which type of agent was the most suitable for different decision stages. The winrates of agent's combinations as seen in Table~\ref{table:1} imply that the optimal mixed strategy is to employ MCTS (Agent 3 and 4) at Planning, whereas Questing and Defense on Expert (Agent 2). Such configuration of agents is able to win over 95\% of games at medium complexity level. Other setups worth noting (winrate over 90\%) are these, which utilize MCTS at Defense stage. One can note that expert agent has poor performance if used on the Planning stage, however this is not a solid conclusion.

\begin{figure}[htbp]
    \centerline{\includegraphics[width=0.5\textwidth]{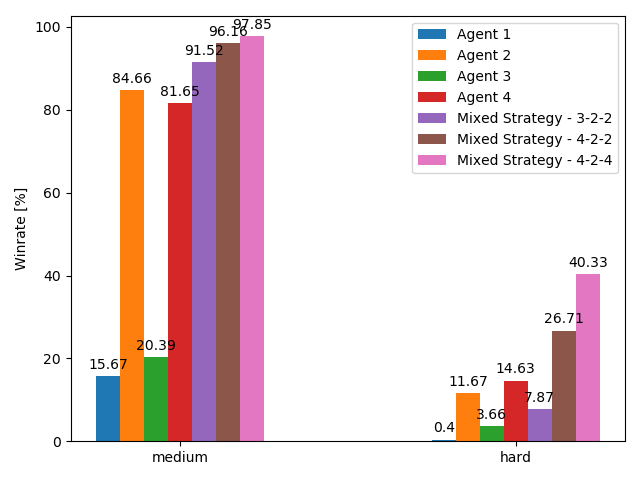}}
    \caption{Final comparison of different AI players winrates for two game complexity levels (medium and hard, number of trials – 10000). First 4
players use the same strategy for each decision stage, the others apply mixtures of strategies.}
    \label{fig:finalcomp}
\end{figure}

Final verification of the developed artificial player had been
done against the game with the highest ("Hard" in Table~\ref{tbl:complexity}) complexity
level and 10 000 game repetitions for more accurate evaluation.
Four players using the same strategy at every decision stage
(1‒1‒1, 2‒2‒2, 3‒3‒3, 4‒4‒4) and the top three mixed strategies
from Table~\ref{table:1} were compared. Results are presented in Fig.~\ref{fig:finalcomp}.
It is clear that level 'hard' is a great challenge for every agent.
Winrate of 0.4\% proves that there is no reason for playing the
game at the hard level using Random Agent. Moreover, the
difference between Agents 3 and 4, shows that implementing
MCTS algorithm with expert knowledge at the Expansion stage is
worth an effort. However, major progress is observed for
the mixed strategies. The undoubted winner was a player with
a strategy agent4 – agent2 – agent4, who defeated all the others
with almost double the lead.

It should be remembered that even at the 'hard' level, the simulated game contains simplifications. In future studies, it is planned to validate the developed methodology on the complete game simulator.

\section{Conclusions}
"The Lord of the Rings" is a popular multi-stage card game with a high degree of randomness, which poses a serious challenge to computational intelligence methods. Although there are studies on similar card games\cite{b5}\cite{Ward2009} \cite{Zhang2017}, the case of LoTR has not previously been analyzed in detail in the scientific literature.

Developed AI agent based on the MCTS algorithms can achieve a significantly higher winrate than an expert, rule-based player. The presented method takes a separate analysis of each of the decision-making stages in the game. Numerical experiments have shown that different methods in different stages allow increasing the overall winning rate.

Another main conclusion is that the inclusion of expert knowledge significantly improves the results of the method. In the proposed solution, additional domain knowledge has been used to reduce the number of analyzed actions at the expansion step of the MCTS but also improve the efficiency of the playouts.

The MCTS algorithm is known as a universal and powerful tool but with high computational requirements. The measured time of simulations confirmed that implemented extensions are not deteriorating their performance.

Future efforts will be directed towards the development of the game simulator without any simplification, as well as the use of other methods of computational intelligence, such as reinforcement learning, to create agents comparable to the developed optimal MCTS player.

\end{document}